\documentclass[11pt,a4paper]{article}
\usepackage[hyperref]{acl2018}
\usepackage{times}
\usepackage{latexsym}
\usepackage{graphicx}
\usepackage{url}
\usepackage[normalem]{ulem}
\usepackage{todonotes}

\usepackage{color}
\usepackage[linesnumbered,ruled]{algorithm2e}
\aclfinalcopy % Uncomment this line for the final submission
\title{Investigating Biases in Textual Entailment Datasets}

\author{
    Shawn Tan \qquad Yikang Shen \qquad
    Chin-wei Huang \qquad Aaron Courville \\
    {\tt \{jing.shan.shawn.tan, yi-kang.shen,
           chin-wei.huang, aaron.courville\} }\\
    {\tt @umontreal.ca}
}

\date{}

\begin{document}
\maketitle

\begin{abstract}
The ability to understand logical relationships between sentences is an important task in language understanding.
To aid in progress for this task, researchers have collected datasets for machine learning and evaluation of current systems.
However, like in the crowdsourced Visual Question Answering (VQA) task, some biases in the data inevitably occur.
In our experiments, we find that performing classification on just the hypotheses on the SNLI dataset yields an accuracy of 64\%.
We analyze the bias extent in the SNLI and the MultiNLI dataset, discuss its implication, and propose a simple method to reduce the biases in the datasets.

\end{abstract}

\section{Introduction}
Natural Language Inference (NLI) is an important task for natural language understanding \cite{maccartney2009extended}.
It involves discerning if a natural language sentence $h$ can reasonably be inferred from an originating sentence $p$.
To this end, several datasets have been collected for the evaluation of a system's ability to detect such relationships between sentences \cite{marelli2014sick,young2014image,bowman2015large,williams2017broad}.
These datasets evaluate models for the task of Recognizing Textual Entailment (RTE), and \citet{bowman2015large} introduced  the Standford Natural Language Inference (SNLI) dataset, a much larger dataset than before, boasting $\sim$500K examples that are crowdsourced with specific constraints.
Since its introduction, there have been numerous proposals for different models to perform this task \cite{chen2017enhanced,gong2017natural}.
Later, a dataset for RTE over a broader set of domains was introduced in \citet{williams2017broad}, the MultiNLI dataset.

Recently, though, in the Visual Question Answering (VQA) dataset \cite{antol2015vqa}, biases due to human predispositions  when generating related questions for images were found.
As an example, one can attain a 68\% accuracy when answering ``yes" to all binary questions in VQA \cite{zhang2016yin}.
This is not only a problem during evaluation, but also results in statistical learning algorithms picking up superficial correlations in the training set, if such biases exist there as well.

Do the SNLI and MultiNLI datasets contain the same type of human biases?
If they do, do current state-of-the-art models for RTE rely too heavily on them, and are there ways to modify the current dataset to correct for it?
%We have to be certain that superficial correlations found in the training data and test data are not exploited to give overly optimistic results during evaluation, resulting a false impression that progress is being made for RTE.
In this paper, we set out to analyse SNLI and MultiNLI, specifically looking for signs of similar types of biases introduced through the data collection mechanism.
We also propose a simple heuristic to try to correct for correlations in superficial aspects of the data, hoping to stir discussion and inspire future work in this direction. 

%Investigating the SNLI dataset, we find biases as well.
%Labels for the RTE task can be predicted from just the hypothesis alone with 68\% accuracy.

%AMT widely used, but if not carefully used results in crappy dataset.

%Working on SNLI

%Found that data is shitty

\section{Related Work}
In the SNLI dataset \cite{bowman2015large}, Amazon Mechanical Turk was used to crowdsource data collection.
In each task, a worker was presented with a premise $p$, and asked to write three hypotheses: \textit{contradictory}, \textit{entailing} and \textit{neutral} sentences.
The premises were obtained from the Flickr30k corpus \cite{young2014image}, and contained $\sim$160K captions.
Additionally, there was a validation step to ensure that four other workers agreed that the written sentence corresponded to the label.
%\todo[inline]{\citet{williams2017broad} MultiNLI bigger than SNLI.}
Similarly, the VQA dataset \cite{antol2015vqa} also crowdsourced questions from Amazon Mechanical Turk.
Workers were asked to provide questions given an image that they believed a ``smart robot'' would have trouble answering. 
However, \citet{zhang2016yin} revealed problems with the VQA dataset related to biases in the questions, including as discussed in the introduction, a bias toward affirmative answers to yes/no questions.
\citet{zhang2016yin} suggest a solution to the affirmation bias by using crowdsourced clipart to generate a dataset where every question has two complementary scenes with opposite answers, effectively ``debiasing" the dataset.
\citet{goyal2017making} has a similar goal, but instead of generating synthetic images, it attempts to identify another image that results in a different answer. This effort was again made possible by additional reliance on crowdsourcing. Another way to sidestep the problem of biased training and test sets is to incorporate debiasing directly into the model. For example,
\citet{agrawal2017don} adapted the design of the architecture of the model explicitly to avoid learning the data bias.

\citet{gururangan2018annotation} and \citet{poliak2018hypothesis} also independently discover such biases in the dataset. 
\citet{gururangan2018annotation} categorized the test set into different levels of difficulty that help evaluate the performance of the model, and \citet{poliak2018hypothesis} emphasized that the statistical irregularity in the hypothesis alone allows the model to achieve NLI without actual pairwise reasoning.
In our work, we reproduce the hypothesis-only results on SNLI, and also try to analyse the dependence on the hypothesis for a model trained for the RTE task.
We also perform a bigram analysis on the training and test set, and propose a simple way to prune the training set based on the bigram distribution.

\section{Analysis}
\subsection{Classification on Hypothesis Only}

\begin{table}[t]
\begin{center}
\begin{tabular}{|l|c|c|}
\hline \bf Dataset & \bf $h$-only  \\
\hline
\textsc{SNLI}    & 64\%    \\
\textsc{MultiNLI}  & 51\%    \\
\hline
\end{tabular}
\end{center}
\caption{Results from using only the hypothesis for classification.}
\label{table:hyponly_table} 
\end{table}

In an effort to probe the bias within SNLI and MultiNLI, we attempt to trained a textual entailment classifier to predict the  \textit{contradictory}, \textit{entailing} and \textit{neutral} labels \emph{from only the hypothesis}.
Intuitively, this should result in almost equal probabilities for each class (assuming balanced classes), for without a premise for comparison, above chance performance should not be possible. 
However, a simple RNN classifier (which we refer to as the $h$-only model) results in a 64\% accuracy on the test set, nearly two times higher than a baseline chance prediction \footnote{The same test was not carried out for premise-labels because there are (approximately) balanced triplets of labels for each premise. Thus, by construction, there should be little or no bias of this type for the premise.}.
\citet{poliak2018hypothesis} further investigates this issue with a more comprehensive study over a wider range of corpora.
This suggests that there are correlations that exist in the training set that can be exploited during test time.
We will further discuss the implications of this in Section \ref{sec:conclusion}.

MultiNLI has multiple genres of data (Fiction, Telephone, Travel, etc.) and they split their development set into two: the \texttt{matched} development set consists of test examples that come from 5 of the genres that are also seen in the training set, while the \texttt{mismatched} development set contains examples from unseen genres.
Running the same experiment on MultiNLI, the same hypothesis-only classfier achieves a 51\% accuracy on the mismatched dataset.
%While still higher than random, the result suggests that the MultiNLI training/test partition has a bigger difference in terms of distribution than that of the SNLI training/test partition.
This may be because the MultiNLI dataset has less superficial correlations that the classifier is able exploit.

\subsection{Testing Hypothesis Dependence for NLI Models}

\begin{table}[t]
\begin{center}
\begin{tabular}{|l|c|c|}

\hline \bf Model Type & \bf RTE & \bf Permuted \\
\hline
\textsc{ESIM}    & 88\%  & 40.5\%  \\
\textsc{LSTM}  & 70\%  & 50\%  \\
\hline
\end{tabular}
\end{center}

\caption{Results from permuting premises. LSTM refers to the sentence-embedding method that use LSTM cell as encoder.}
\label{table:permute_results} 
\end{table}

As one of the reasons for the NLI task was for the learning of sentence representations, we also trained an LSTM sentence-embedding encoder.
The idea was to compare the performance between a model that uses a fixed-length sentence embedding and one that  tries to model interactions between hidden states of an RNN (ESIM and DIIN fit into this category).
Because sentence embedding models do not force the `interaction' between the two inputs, we believe that the sentence embedding models may be more prone to learning these superficial correlations.

The experiment attempts to test sentence-embedding models for their reliance on the hypothesis for classification.
During testing, we shuffle the premises so that they do not correspond to the right hypotheses.
The sentence-embedding models that we trained achieved 70\% accuracy when trained on the full dataset while under the shuffled premise test, they achieved an accuracy of 50\%.
In comparison, the ESIM model achieved a 40.5\% accuracy in this setting.
This suggests that the model still uses some of the correlations found in the hypothesis, otherwise this experiment should result in an $\sim$33\% accuracy.
The results hint that a sentence embedding model has a stronger reliance on the hypothesis and, therefore, the biases in the dataset.

\subsection{Bigrams}

\begin{figure}
\begin{center}
\includegraphics[width=\linewidth]{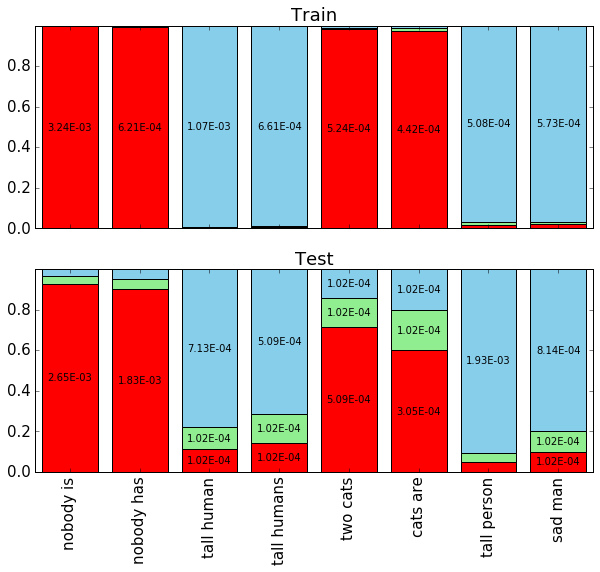}
\end{center}
\caption{\label{fig:snli_ratios}
The top most informative bigrams in the SNLI dataset.
{\color{red} Red} represents proportion of contradiction labels,
{\color{blue} Blue} for neutral,
and {\color{green} Green} for entailment.
Numbers on the bars represent the proportion of the bigram in the dataset (A bar labeled with 0.5 means that portion of the bigram constitutes half of that partition of the dataset). 
}
\end{figure}

\begin{figure}
\begin{center}
\includegraphics[width=\linewidth]{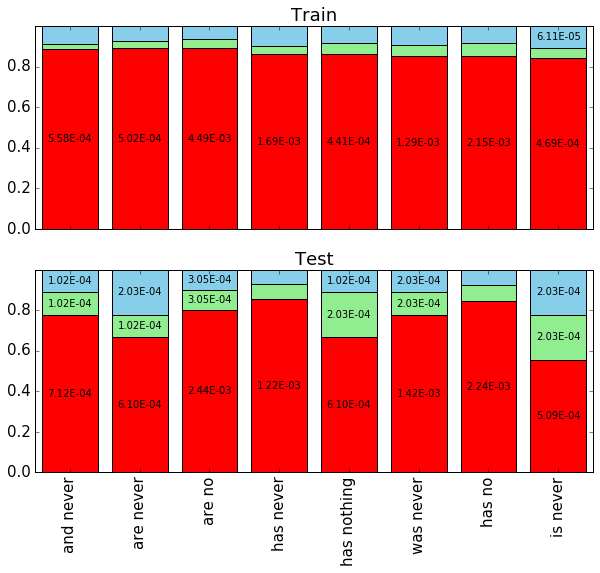}
\end{center}
\caption{\label{fig:mnli_ratios} The top most informative bigrams in the MultiNLI dataset. The color legend is identical to Fig.~\ref{fig:snli_ratios}}
\end{figure}
We analyze the most informative bigrams that are in the SNLI training set.
Specifically, we count the bigrams $w$ in each class $c$, and calculate $p(c \mid w)$ for each $w$ that occurs more than $200$ times, then applying Laplace smoothing with $\alpha=1$ to the counts before normalizing by the total counts.
We then rank them in order of increasing entropy,
 $H(c \mid w) = - \sum_c p(c \mid w) \log p(c \mid w).$
The distributions with least entropy are shown in Figure \ref{fig:snli_ratios} for SNLI and \ref{fig:mnli_ratios} for MultiNLI.
This is then compared to their proportions seen in the test set, in order to get an idea of the frequency of their occurrences in both partitions.

In the test set, their ratios across classes appears to be relatively similar to the training set.
But because of the size of the test set in comparison to the training set ($\sim$50 times smaller), and coupled with smoothing, the distributions are more uniform.
For SNLI, we find that the informative bigrams make up the long-tail of the bigram distributions, but many of them are predictive of the labels.
MultiNLI also has many low frequency bigrams that are preferentially predictive of contradiction.
These bigrams tend to correspond to negative notions (e.g. \textbf{never}, \textbf{no}, \textbf{nothing}).
In comparison with SNLI, the odds of the highest information bigram in SNLI, \textbf{nobody is}, predicts for contradictions 222:1.
For MultiNLI, \textbf{and never} predicts for contradiction 8:1.

\begin{table}
\begin{center}
\small
\begin{tabular}{|l|l|p{0.6\linewidth}|}
\hline \bf Label & & \bf Examples \\ \hline
Contradiction & P & Black man in a nice suite that matches the rest of the choir he's singing with near a piano. \\
              & H & \textbf{nobody is} singing \\
\hline
Neutral & P & An excited, smiling woman stands at a red railing as she holds a boombox to one side. \\
           & H & A \textbf{tall human} stanindg. \\
\hline
Entailment & P &	A group of people are walking across the street. \\
        & H & \textbf{some humans} walking
\\
\hline
\end{tabular}
\end{center}
\caption{\label{table:lazy_examples} Examples of top bigram occurrences for each label in SNLI.}
\end{table}

Picking examples that contain these bigrams from SNLI, we can understand why they were repeatedly used to generate hypotheses for those classes (Table \ref{table:lazy_examples}).
The most informative bigram, \textbf{nobody is/has} was often used when the premise describes someone performing a task.
The turker simply has to substitute ``nobody'' into the sentence in order to make the sentence a contradiction.
The bigram \textbf{tall human} was used to inject an additional detail in the sentence, while at the same time being less detailed about the person in question, resulting in a neutral hypothesis.
To create an entailment sentence, using \textbf{some humans} resulted in a sentence that could be entailed from the premise, but removed details about what type of human it was.
We also notice that there are fewer bigrams that are preferential to entailments, in both SNLI and MultiNLI. 
One simple reason for this is that one just needs to remove details from the premise, instead of adding extra information, in order to generate an entailed sentence.
Thus, it is relatively easy to construct entailed sentences without incurring significant bias.

\section{Correcting SNLI via dataset pruning}
If we know that the probability on all the classes should be almost equal given only $h$, then ideally each $h$ should have an equal number of pairings with every class.
%\citet{zhang2016yin} modified the dataset by using synthetic images to provide a balanced dataset given each question.
In an attempt to reduce the bias of SNLI, we prune the training dataset using the features of the hypothesis. % for this purpose.
Pruning the dataset to balance the feature occurrence should result in a distribution shift between the train and test set.
If the model has learned to do logical inference, the bias in the test set should make relatively little difference.
%assuming all of them have been assigned the right logical entailment label.
\subsection{Greedy Pruning}

\begin{algorithm}[t]
 \caption{The classifier greedy removal algorithm.} \label{algo:greedy}
    \SetKwInOut{Input}{Input}
    \SetKwInOut{Output}{Output}

    \underline{function \textsc{PruneDataset}}$(\mathcal{D}, \lambda)$\\
    \Input{The original dataset $\mathcal{D}$}
    \Input{Proportion of dataset to prune $\lambda \in (0, 1)$}
    \Output{$\mathcal{D}'$}
    $\hat{p} \leftarrow \textsc{TrainClassifier}(\mathcal{D})$\;
    $\mathcal{D}' \leftarrow\mathcal{D}$\;
    \For{$i \leftarrow 1 \dots \lambda |\mathcal{D}|$}{
        $(h', c') \leftarrow
         \mathop{\arg \min}\limits_{(h, c) \in \mathcal{D'} }
         -\log \hat{p}(c|h) $\;
        $\mathcal{D}' \leftarrow \mathcal{D}' \setminus \{(h', c')\}$\;
        $\hat{p} \leftarrow \textsc{TrainClassifier}(\mathcal{D}')$\;
    }
  
\end{algorithm}

In our approach to re-balancing the training dataset, we rely on iteratively retraining a simple classifier.
Since we know that bigrams in the hypothesis are predictive of the labels, we use bigrams as features for a Naive Bayes classifier.

Every time we remove an instance from the dataset, the most informative features may change (the frequencies of other bigrams present in that instance are affected).
If we remove data instances without taking this shift into account, a new set of instances would become the most informative.
To deal with this, a classifier should be retrained for every iteration of the pruning.
The reason Naive Bayes was used for pruning was because it was easy to retrain to optimality given the original dataset by simply subtracting the counts.

Using the predictions of the classifier on the the training set, we score the instances in the dataset by their cross-entropy.
We then remove the instance with the lowest cross entropy and update the classifier accordingly.

Our goal is to ensure that the distribution of classes for each bigram is balanced.
However, since each instance contains several bigrams, and we want to remove as few instances as possible (to maximize diversity), we score each instance with how predictive the bag of bigrams are together.
A Naive Bayes model was chosen because it was easy to update the classifier at every iteration by subtracting bigram counts from the model.

\begin{figure}
\begin{center}
\includegraphics[width=\linewidth]{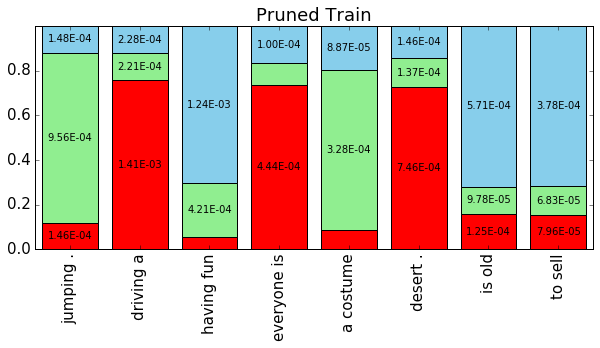}
\end{center}
\caption{\label{fig:pruned_snli_ratios} The top most informative bigrams in the Pruned SNLI dataset.}
\end{figure}

Algorithm \ref{algo:greedy} lists the pseudo-code of the method. 
Figure \ref{fig:pruned_snli_ratios} shows the most informative bigrams on the pruned version of the dataset.
As compared to the uncorrected SNLI, the top 8 most informative bigrams are less predictive of the class label.

\begin{table}[t]
\begin{center}
\begin{tabular}{|l|l|l|l|}
\hline \bf Method & \bf $h$ only & \bf Train & \bf Test \\
\hline
\textsc{Original}   & 64\% & 93\% & 88\% \\
\textsc{Random}     & 59\% & 94\% & 87\% \\
\textsc{Greedy}     & 56\% & 81\% & 85\% \\
\hline
\end{tabular}
\end{center}

\caption{Results from training on the RTE task. \textbf{$h$ only} uses only the hypothesis for classification. \textbf{Train} and \textbf{Test} are the results from training the ESIM model \citep{chen2017enhanced} on the various datasets.
\textsc{Random} refers to the dataset in which we uniformly remove 20\% of the dataset, and \textsc{Greedy} refers to using our greedy pruning method to remove 20\% of the instances from the dataset.
}
\label{table:classifier_results} 
\end{table}
%\textsc{Original}   & 64\% & 93\% & 88\% \\
%\textsc{Random}     & 59\% & 94\% & 87\% & 61\%\\
%\textsc{Greedy}     & 56\% & 81\% & 85\% & \\
We perform the RTE task using our $h$-only model on hypotheses alone, and the ESIM model on the hypothesis-premise pairs. The ESIM model was used in this analysis because it is one of the models with state-of-the-art results, and the ease of working with the code-base.

To measure how the pruning of the training set affects the classification task, we compare training on the pruned dataset with training on the full, original dataset, and a uniformly randomly pruned dataset as a control to calibrate the effect of smaller trianing set size on generalisation.
We refer to these as the \textsc{Original} and \textsc{Random} strategies respectively, and the strategy we propose as \textsc{Greedy}.
The result is presented in Table \ref{table:classifier_results}.

Interestingly, using the \textsc{Random} strategy, the model performs the same  on the RTE tasks.
However, running it on the $h$-only classifier resulted in a lower accuracy.
It is possible that sufficient numbers of the label-predictive bigrams were removed that the classifier is less able to exploit these for classification.
More surprisingly, our removal method, while resulting in a 3\% drop on the test set, also results in a lower accuracy on the training set.
We believe this is due to the training set becoming a much harder dataset on which to train, with fewer statistical correlations between hypothesis and label. 
Also, higher performance on hypothesis alone correlates with higher performance on both hypothesis and premise. 
This indicates the reported measurements on performance of the state-of-the-art models are overestimated since the class label should be marginally independent of any single sentence alone.

%\todo[inline]{Check out the interesting table, some points:

%Randomly removing the data from the dataset doesn't really help. 

%Interesting fact about greedy method, works better on test set than on training. Overall works worse.}
\section{Discussion \& Conclusion}\label{sec:conclusion}
%The prior work on debiasing VQA had two important takeaways that we apply in our work.

The NLI datasets were created in order to train models that learn to perform RTE, with the intention of learning good semantic representations for the task.
In this paper, we present the biases present in the data, and how they are similar in both the training and test set.
Most statistical learning algorithms will exploit available superficial correlations, and then be evaluated on the test set that is similarly biased.
This results in a score that may not be representative of how well the field is advancing towards true RTE performance.
There are two key takeaways we would like to emphasise:

\paragraph{Train / test split with different distributions for proper benchmarks} 
If the partition is made such that the distributions between train and test are different, any unwanted correlations between the hypotheses and labels in the  training set cannot be exploited during testing.
This effectively prevents the information about the test set from `leaking' into the training data.
What this means is that in order to have a score that reflects the state of the art in the task, we should have differently biased train and test sets.

\paragraph{Conditional independence of the label and hypothesis}
Without the premise, the label should be conditionally independent of the hypothesis, and a model that performs RTE should manifest this behaviour.
One way to achieve this is to ensure that the dataset reflect the true dependence of the textual entailment labels on the relationship between premise and hypothesis, not on a set of marginal features of the hypothesis.
Alternative methods are possible, including losses that enforce conditional independence in the model.

In this paper, we proposed a simple method based on bigrams.
In pruning the training set and keeping the test set the same, we are attempting to change the distribution in the train and test partitions, and reduce marginal features of the hypothesis so that the learning algorithm does not exploit the superficial correlations.
These properties should be kept in mind when training a model on a dataset, or when assessing collected data that is being curated for a dataset.
%Using machine learning methods to solve a task requires either that the training dataset represents a relatively unbiased sample of the true distribution for that task, or, at the least that the training distribution allows the model to learn the discriminative features that allow  the model to generalize to the task distribution. 
%For tasks like VQA and RTE, it appears to be difficult to construct a dataset free of bias.
%This suggests that we should focus on having the latter instead. However, we still want to be able to learn a good model from the training data, so just having different distributions is not enough

From our analysis, we believe MultiNLI to have fewer issues with bias compared to SNLI.
If SNLI is still preferred, some preprocessing should be performed in order to account for the problems we mentioned.
%Another avenue of combating such problems that was left unmentioned in our paper was incorporating the de-biasing into the model, or the loss. We leave this for future work.
We hope that the issues we have raised will help researchers to better diagnose and analyse their results.
%Perhaps there is no way to get unbiased datasets for such tasks.

%Best we can do may be to partition or collect data such that the train/test split is explicitly constructed to be different.

%Balancing the dataset is harder, so doesn't seem to be done.
\bibliography{acl2018}
\bibliographystyle{acl_natbib}

\end{document}